\documentclass[runningheads]{llncs}
\usepackage{amsmath, amssymb, amsfonts}
\usepackage{url}
\usepackage{xcolor}
\usepackage{graphicx}
\usepackage{verbatim}
\usepackage{latexsym}
\usepackage{setspace}
\renewcommand{\vec}[1]{\mathbf{#1}}
\newtheorem{prop}{Proposition}

\DeclareMathOperator*{\argmax}{arg\,max}

\graphicspath{{figures/}}

\begin{document}
\title{Jacobian Ensembles Improve Robustness Trade-offs to Adversarial Attacks\thanks{Kenneth T. Co is supported in part by the DataSpartan research grant DSRD201801.}}
%
%
\author{Kenneth T. Co\inst{1,2}\orcidID{0000-0003-2766-7326} \and
David Martinez-Rego\inst{2}\orcidID{0000-0003-1809-1169} \and 
Zhongyuan Hau\inst{1}\and
Emil C. Lupu\inst{1}\orcidID{0000-0002-2844-3917}}
\authorrunning{K. Co et al.}
%
\institute{Imperial College London, London SW7 2AZ, United Kingdom
\email{\{k.co,zy.hau17,e.c.lupu\}@imperial.ac.uk}\\
\and DataSpartan, London EC2Y 9ST, United Kingdom\\
\email{david@dataspartan.com}}
\maketitle       
\begin{abstract}
Deep neural networks have become an integral part of our software infrastructure and are being deployed in many widely-used and safety-critical applications. However, their integration into many systems also brings with it the vulnerability to test time attacks in the form of Universal Adversarial Perturbations (UAPs). UAPs are a class of perturbations that when applied to \textit{any input} causes model misclassification. Although there is an ongoing effort to defend models against these adversarial attacks, it is often difficult to reconcile the trade-offs in model accuracy and robustness to adversarial attacks. Jacobian regularization has been shown to improve the robustness of models against UAPs, whilst model ensembles have been widely adopted to improve both predictive performance and model robustness. In this work, we propose a novel approach, Jacobian Ensembles -- a combination of Jacobian regularization and model ensembles to significantly increase the robustness against UAPs whilst maintaining or improving model accuracy. Our results show that Jacobian Ensembles achieves previously unseen levels of accuracy and robustness, greatly improving over previous methods that tend to skew towards only either accuracy or robustness. 

\end{abstract}

\section{Introduction}
Deep neural networks (DNNs) have achieved widespread use in many applications including image classification \cite{krizhevsky2012imagenet}, real-time object detection \cite{redmon2016you}, and speech recognition \cite{hinton2012deep}. Despite these advances, there is an increasing recognition that DNNs are exposed to systemic vulnerabilities in the form of Universal Adversarial Perturbations (UAPs): where a single adversarial perturbation causes a model to misclassify a large set of inputs \cite{moosavi2017universal}. Thus, it is important to ensure that neural networks are robust to such devastating attacks whilst still maintaining their state-of-art accuracy on benign datasets.

UAP present a systemic risk, as they enable practical and physically realizable adversarial attacks. They have been demonstrated in many widely-used and safety-critical applications such as camera-based computer vision \cite{eykholt2018physical,eykholt2018robust,brown2017adversarial,matachana2020robustness} and LiDAR-based object detection \cite{hau2021object,hau2021shadow}. UAPs have also been shown to facilitate realistic attacks in both the physical \cite{thys2019fooling} and digital \cite{tramer2019adversarial} domains. In some cases, UAPs also enable very resource-efficient black-box attacks on DNNs \cite{co2019procedural,co2019sensitivity}.

An insufficiently studied aspect of existing defenses against UAPs is the trade-off between clean accuracy, or the model's performance on a benign dataset, and its robustness to adversarial attacks. Indeed, a model with increased robustness to UAPs is desirable, but reduced clean accuracy could translate to reduced utility and additional financial or security costs depending on the application. Existing defenses primarily consider robustness to adversarial attacks, but neglect the cost it incurs on the model's performance for the original task. For example, defenses like adversarial training or too much regularization improve robustness but greatly reduce clean accuracy \cite{co2021jacobian,hoffman2019robust,roth2020adversarial}.

\textit{Jacobian regularization} (JR) has previously been shown to improve robustness against UAPs. However, JR can damage clean accuracy for large amounts of regularization \cite{co2021jacobian,hoffman2019robust,roth2020adversarial}. \textit{Model ensembles} on the other hand has widely been shown to achieve better classification performance and stability than a single (best) classifier \cite{kuncheva2014combining,zhou2012ensemble}. Ensembles are created by combining the outputs of multiple base learners to generate an improved prediction.

In this work, we propose combining JR and model ensembles during training to create \textit{Jacobian Ensembles}. JR is used to drastically improve the model's robustness to UAPs while the ensemble methods stabilize the model's predictions and improve its clean accuracy. JR and model ensembles individually each have been shown to improve UAP robustness but at some cost to clean accuracy \cite{co2019universal,co2021jacobian}. We show that Jacobian Ensembles greatly improve on the accuracy-robustness trade-off when compared to either JR or model ensembles individually. First, we theoretically show that increasing the number of base learners in a model ensemble improves the expected robustness of classifiers. Then, we empirically verify our theoretical findings by applying JR with popular ensemble methods bagging \cite{breiman1996bagging}, snapshot ensembles \cite{huang2017snapshot}, soft voting \cite{zhou2012ensemble} to DNNs trained on the popular benchmark datasets: MNIST \cite{lecun1998gradient}, Fashion-MNIST \cite{xiao2017fashion} and then evaluating their robustness against UAPs.

To summarize, we make the following contributions:
\begin{itemize}
    \item We derive theoretical formulations for robustness of ensemble methods and show that the robustness to UAPs increases monotonically with the number of base learners.
    \item We empirically verify our theoretical results and show that Jacobian Ensembles, a combination of Jacobian regularization and ensembles, achieves the best accuracy-robustness trade-off as measured by a combined weighted accuracy metric.
\end{itemize}
\section{Background}

\subsection{Universal Adversarial Perturbations}
Let $f: \mathcal{X} \subset  \mathbb{R}^D \to \mathbb{R}^C$ be logits of a piece-wise linear classifier with input $\vec{x} \in \mathcal{X}$. We, define $F(\vec{x}) = \argmax(f(\vec{x}))$ to be the output of this classifier and write $\tau(\vec{x)}$ as the true class label of an input $\vec{x}$. \textbf{Universal Adversarial Perturbations (UAP)} are perturbations $\delta \in \mathbb{R}^n$ to the data that satisfy $F(\vec{x} + \delta) \neq \tau(\vec{x})$ for sufficiently many $\vec{x} \in \mathcal{X}$ where $\Vert \delta \Vert_p < \varepsilon$. The latter condition $\Vert \delta \Vert_p < \varepsilon$ constrains the magnitude of the perturbation and is often some $\ell_p$-norm and small $\varepsilon > 0$ \cite{moosavi2017universal}.
Given a classifier $f$, UAPs are generated by maximizing the loss $\sum_i \mathcal{L}_f(\vec{x}_i + \delta)$ with an iterative stochastic gradient descent algorithm \cite{co2019universal,shafahi2018universal} where $\mathcal{L}$ is the model's training loss, $\{\vec{x}_i\}$ are batches of inputs, and $\delta$ are small perturbations that satisfy $\Vert \delta \Vert_p < \varepsilon$.

\subsection{Model Ensembles}
An ensemble consists of combining multiple classifiers (base learners) to obtain a resulting ensemble that has better accuracy or predictive performance on aggregate than any individual base learner. In practice, it is widely accepted that combining multiple classifiers can achieve better classification performance than a single ``best'' classifier \cite{kuncheva2014combining,zhou2012ensemble}. Ensembles are typically generated in two ways: sequentially and in parallel. After generating the base learners the combination of their outputs is taken rather than choosing a single ``best'' learner \cite{zhou2012ensemble}.

In this work, we will analyze ensemble methods that aggregate their base learners in a convex combination. Formally, we define an ensemble $\mathcal{F}$ as a convex combination of $M$ base learners $f_i$: $\mathcal{F}(x) = \sum_{i = 1}^M c_i f_i(x)$ where $\sum_{i = 1}^M c_i = 1$ and $0 < c_i < 1$, $\forall i$. This is typical as many ensemble methods aggregate their methods via averaging or a similar form of weighted sum \cite{zhou2012ensemble}. Note however that this will exclude some boosting algorithms such as AdaBoost that are typically not convex combinations \cite{freund1999short}.

Popular algorithms like \textit{Bagging} \cite{breiman1996bagging} and newer methods like \textit{Snapshot Ensembles} \cite{huang2017snapshot} take the average of their base learners. Other methods like \textit{Soft Voting} \cite{zhou2012ensemble} use a convex combination of their model outputs to vote. We refer the reader to each ensemble methods' corresponding paper for further details on how the base learners are generated. For this work, we will consider Bagging, Snapshot Ensembles, and Soft Voting.

\subsection{Jacobian Regularization}
Let $f(\vec{x})$ be the logit output of the classifier for input $\vec{x}$, we write $\vec{J}_{f}(\vec{x})$ to denote the input-output Jacobian of $f$ at $\vec{x}$. To train models with Jacobian regularization (JR) \cite{hoffman2019robust,co2021jacobian}, the following joint loss is optimized:
\begin{equation}\label{eq-JR}
    \mathcal{L}_{\text{joint}}(\vec{\theta}) =  \mathcal{L}_{\text{train}}(\{\vec{x}_i, \vec{y}_i\}_i, \theta) + \frac{\lambda_{\text{JR}}}{2} \left( \frac{1}{B} \sum_i \Vert \vec{J}(\vec{x}_i) \Vert_F^2 \right)
\end{equation}
where $\theta$ represent the parameters of the model, $\mathcal{L}_{\text{train}}$ is the standard cross-entropy training loss, $\{\vec{x}_i, \vec{y}_i\}$ are input-output pairs from the mini-batch, and $B$ is the mini-batch size. This optimization uses a regularization parameter $\lambda_{\text{JR}}$, which allows the adjustment between regularization and classification loss.

The primary idea is to reduce the Frobenius norm of the input-output Jacobian $\Vert \vec{J}(\vec{x}_i) \Vert_F$ to decrease the model's sensitivity to small perturbations such as UAPs. JR shows some promise in improving robustness to UAPs. However, it can often simultaneously decrease the model's clean accuracy \cite{co2021jacobian} especially for large values of $\lambda_{\text{JR}}$.

\section{Bounds on UAP Effectiveness for Model Ensembles}
In this section, we derive theoretical bounds for the expectation and variance of the Frobenius norm of the Jacobian of model ensembles. Note that we only consider ensembles that take a convex combination of their base learners. Similar to \cite{co2021jacobian}, we restrict the Frobenius norm of the Jacobian of a model to improve robustness against UAPs.

We show that using model ensembles result in tighter bounds on the Frobenius norm of the Jacobian, suggesting improved robustness and stability versus a single classifier. Our main result in \textbf{Theorem~\ref{thrm:jac-ens}} shows that increasing the number of base learners decreases both the upper and lower bounds of the expectation and variance for the Frobenius norm of the ensemble's Jacobian.

\begin{prop}\label{prop:jac-ens}
Let $x_i$ be independent random variables drawn from a Normal distribution $x_i \sim \mathcal{N}(\mu, \sigma^2)$ for a fixed mean $\mu$ and variance $\sigma^2$. Define $\overline{x} = \sum_{i = 1}^M c_i x_i$ where $\sum_{i = 1}^M c_i = 1$ and $0 < c_i < 1$, $\forall i$. We then have the following:
    \begin{equation}\label{eq:ens-sigma}
        \frac{\sigma^2}{M} \leq \sum_{i = 1}^M c_i^2 \sigma^2 < \sigma^2
    \end{equation}
\end{prop}
\begin{proof}
    By linearity of Normal distributions: $\overline{x} \sim \mathcal{N}(\mu, \sum_{i = 1}^M c_i^2 \sigma^2)$. We then derive bounds for $\sum_{i = 1}^M c_i^2$ when $M \geq 2$:
        \begin{align*}
        \sum_{i = 1}^M c_i^2 &= \sum_{i = 1}^M c_i^2 + 2 \sum_{i = 1}^M \sum_{j \neq i} c_i c_j - 2 \sum_{i = 1}^M \sum_{j \neq i} c_i c_j\\
        &= \left(\sum_{i = 1}^M c_i\right)^2 - 2 \sum_{i = 1}^M \sum_{j \neq i} c_i c_j\\
        &= 1 - 2 \sum_{i = 1}^M \sum_{j \neq i} c_i c_j
        \end{align*}
    Since $c_i c_j > 0$ for all pairs $i, j$, then it follows that we have the upper bound $\sum_{i = 1}^M c_i^2 < 1$. Note that equality: $\sum_{i = 1}^M c_i^2 = 1$ is only possible in the degenarate case (when $c_i = 1$ for exactly one $i$ and $c_j = 0$ for $i \neq j$).\\
    \noindent For the lower bound, we use Cauchy-Schwarz inequality to get: $$\left( \sum_{i = 1}^M (c_i \cdot 1) \right)^2 \leq \left( \sum_{i = 1}^M c_i^2 \right) \left( \sum_{i = 1}^M 1^2 \right) = \left( \sum_{i = 1}^M c_i^2 \right) \cdot M$$
    The left hand side reduces to 1, so it follows that $\sum_{i = 1}^M c_i^2 \geq \frac{1}{M}$ with equality when $c_i = \frac{1}{M}$ for all $i$. Multiplying all sides with $\sigma^2$ gives the desired bounds.\qed
\end{proof}

Let $\mathcal{F}$ be an ensemble of $M$ base learners $f_i$: $\mathcal{F}(x) = \sum_{i = 1}^M c_i f_i(x)$ where $\sum_{i = 1}^M c_i = 1$ and $0 < c_i < 1$, $\forall i$. Let $\vec{J}_{\mathcal{F}}$ denote the Jacobian of $\mathcal{F}$ and $\vec{J}_i$ the Jacobian of $f_i$ for all $i$. It follows that $\vec{J}_{\mathcal{F}} = \sum_{i = 1}^M c_i \vec{J}_i$.

\begin{theorem}\label{thrm:jac-ens}
Let each matrix $\vec{J}_i \in \mathbb{R}^{C \times D}$ be comprised of the independent random variables $_i a_{pq} \sim \mathcal{N}(\mu, \sigma^2)$, where $_i a_{pq}$ is the element on the $p$-th row and $q$-th column of matrix $\vec{J}_i$. It follows that their convex combination $\vec{J}_{\mathcal{F}}$ satisfies:
\end{theorem}
    \begin{align}
    CD \left( \frac{\sigma^2}{M} + \mu^2 \right) &\leq \text{E}(\Vert \vec{J}_{\mathcal{F}} \Vert_F^2) < \text{E}(\Vert \vec{J}_i \Vert_F^2)\\
    CD \left( \frac{4\mu^2\sigma^2}{M} + \frac{2\sigma^4}{M^2} \right) &\leq \text{Var}(\Vert \vec{J}_{\mathcal{F}} \Vert_F^2) < \text{Var}(\Vert \vec{J}_i \Vert_F^2)
    \end{align}
\begin{proof}
Taking the square of Frobenius norm, we have the following for the Jacobian of a single model:
$$\left\Vert \vec{J}_i \right\Vert_F^2 = \sum_{p = 1}^C \sum_{q = 1}^D |_ia_{pq}|^2$$
The moments of $\left\Vert \vec{J}_i \right\Vert_F$ are proportional to the moments of the random variables $_ia_{pq}^2$. These follow a chi-squared distribution with 1 degree of freedom, and have the expectation and variance:
    \begin{align*}
    \text{E}(_ia_{pq}^2) &= \text{Var}(_ia_{pq}) + [\text{E}(_ia_{pq})]^2\\
    &= \sigma^2 + \mu^2
    \\
    \text{Var}(_ia_{pq}^2) &= \text{E}(_ia_{pq}^4) - [\text{E}(_ia_{pq}^2)]^2\\
    &= \mu^4 + 6 \mu^2 \sigma^2 + 3 \sigma^4 - \mu^4 - 2 \mu^2 \sigma^2 - \sigma^4\\
    &= 4 \mu^2 \sigma^2 + 2 \sigma^4
    \end{align*}
Define $\overline{a}_{pq} = \sum_{i = 1}^M c_i \, _ia_{pq}$, the elements of $\vec{J}_{\mathcal{F}}$. Note that $\overline{a}_{pq} \sim \mathcal{N}(\mu, \sum_{i = 1}^M c_i^2 \sigma^2)$. Thus for the ensemble model's Jacobian, we have:
$$\left\Vert \vec{J}_{\mathcal{F}} \right\Vert_F^2 = \sum_{p = 1}^C \sum_{q = 1}^D |\overline{a}_{pq}|^2$$
It is clear that the moments of $\Vert \vec{J}_{\mathcal{F}} \Vert_F$ are proportional to that of $\overline{a}_{pq}^2$. These random variables follow a chi-squared distribution with $M$ degrees of freedom, and have the following expectation and variance:
    \begin{align*}
    \text{E}(\overline{a}_{pq}^2) &= \sum_{i = 1}^M c_i^2 \sigma^2 + \mu^2\\
    \text{Var}(\overline{a}_{pq}^2) &= 4 \mu^2 \sum_{i = 1}^M c_i^2 \sigma^2 + 2 \left(\sum_{i = 1}^M c_i^2 \sigma^2\right)^2
    \end{align*}
Applying \textbf{Proposition~\ref{prop:jac-ens}}, we then have the following bounds for the expectation and variance for these random variables:
    \begin{align*}
    \frac{\sigma^2}{M} + \mu^2 &\leq \text{E}(\overline{a}_{pq}^2) < \text{E}(_ia_{pq}^2)\\
    \frac{4\mu^2\sigma^2}{M} + \frac{2\sigma^4}{M^2} &\leq \text{Var}(\overline{a}_{pq}^2) < \text{Var}(_ia_{pq}^2)
    \end{align*}
As the random variables are independently drawn, our desired result follows:
    \begin{align*}
    CD \left( \frac{\sigma^2}{M} + \mu^2 \right) &\leq \text{E}(\Vert \vec{J}_{\mathcal{F}} \Vert_F^2) < \text{E}(\Vert \vec{J}_i \Vert_F^2)\\
    CD \left( \frac{4\mu^2\sigma^2}{M} + \frac{2\sigma^4}{M^2} \right) &\leq \text{Var}(\Vert \vec{J}_{\mathcal{F}} \Vert_F^2) < \text{Var}(\Vert \vec{J}_i \Vert_F^2)\quad\qed
    \end{align*}
\end{proof}

\vspace{3pt}\noindent\textbf{Conclusion.} These are proportional to the expectation and variance of the Frobenius norms of our Jacobian matrices, so we can derive the following conclusions in this scenario. Ensembles decrease both the expected value and variance of the Jacobian's Frobenius norms when compared to that of a single model's. As $M$ increases, the lower bounds of both the expectation and variance decreases.

Averaging is one of the most common methods for aggregating base learner outputs in an ensemble \cite{kuncheva2014combining,zhou2012ensemble}, so it is important to consider this case. When the ensemble is done via averaging, i.e. $c_i = \frac{1}{M}$ for all $i$, this achieves the equality condition for the lower bounds of both $\text{E}(\overline{a}_{pq}^2)$ and $\text{Var}(\overline{a}_{pq}^2)$. Therefore, increasing the number of models in the ensemble \textit{strictly decreases} the expectation and variance of the ensemble's Jacobian's norm $\Vert \vec{J}_{\mathcal{F}} \Vert_F$.

This theoretical result gives us the motivation on how model ensembles also improve the stability of models and thus their robustness to UAPs. We show this by deriving the above bounds on the Frobenius norm on their Jacobian. As model ensembles have also been shown to have improved performance over a single classifier, this makes it an ideal candidate for improving both model accuracy and robustness. In the next section, we explore the robustness of model ensembles and verify our theory with empirical results.

\section{Experiments with Jacobian Ensembles}

\subsection{Experimental Setup}

\textbf{Jacobian Ensembles.} To apply Jacobian Ensembles, we only need to include the Jacobian regularization as described in Eq.~\ref{eq-JR} to the joint loss of standard ensemble methods. The Jacobian regularization parameter $\lambda_{\text{JR}}$ is tested for values between 0 and 2: where the resulting models manage to maintain good accuracy as informed by previous work \cite{co2021jacobian}.

We evaluate the following ensemble methods: \textit{Bagging} \cite{breiman1996bagging}, \textit{Snapshot Ensembles} \cite{huang2017snapshot}, and \textit{Soft Voting} \cite{zhou2012ensemble}. For these, we evaluate all experiments with ensembles trained on 1, 3, 6, and 9 base learners. Effectively, one base learner is similar to using no ensemble method at all.

\vspace{3pt}\noindent\textbf{Models \& Datasets.} We use the MNIST \cite{lecun1998gradient} and Fashion-MNIST \cite{xiao2017fashion} datasets. These are popular image classification benchmarks, each with 10 classes, and 28 by 28 pixel images whose their pixel values range from 0 to 1. For the DNN architecture, we use a version of LeNet-5 \cite{lecun1998gradient,hoffman2019robust}, which we refer to as LeNet.

\vspace{3pt}\noindent\textbf{UAP Attacks.} We evaluate the robustness of these models to UAPs generated via iterative stochastic gradient descent with 100 iterations and a batch size of 200. Perturbations are applied under $\ell_{\infty}$-norm constraints. The $\varepsilon$ we consider in our attacks for this norm are from 0.10 to 0.25, this perturbation magnitude is equivalent to 10\%-25\% of the maximum total possible change in pixel values. UAPs are generated over 50 different random seeds, and we report UAPs with the highest attack success rate, as this would represent the worst-case scenario.

\begin{figure}[ht]
\includegraphics[width=\textwidth]{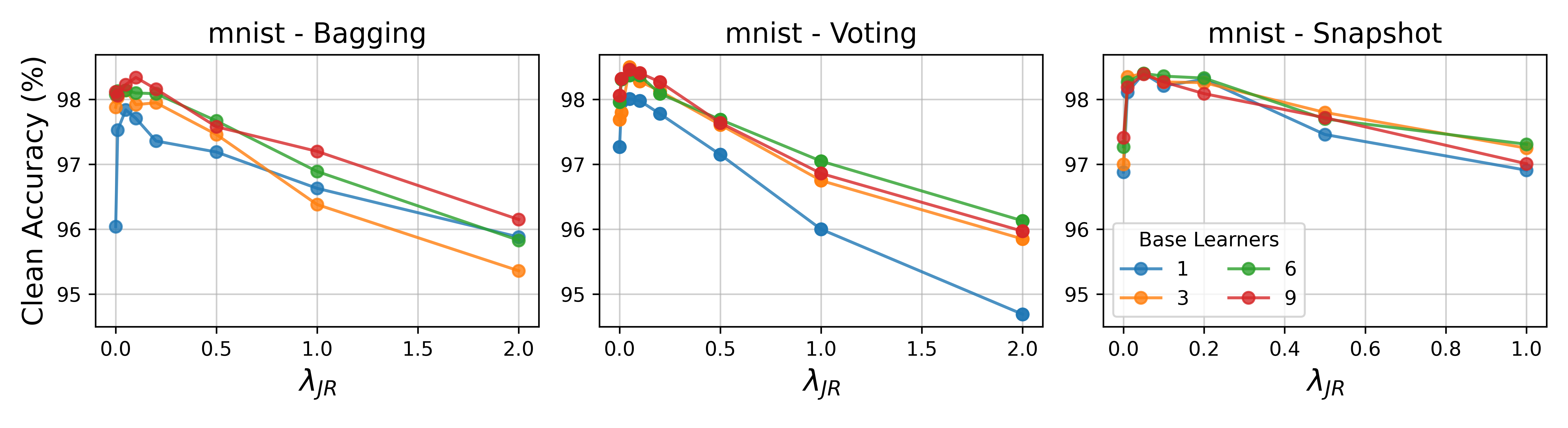}\\
\includegraphics[width=\textwidth]{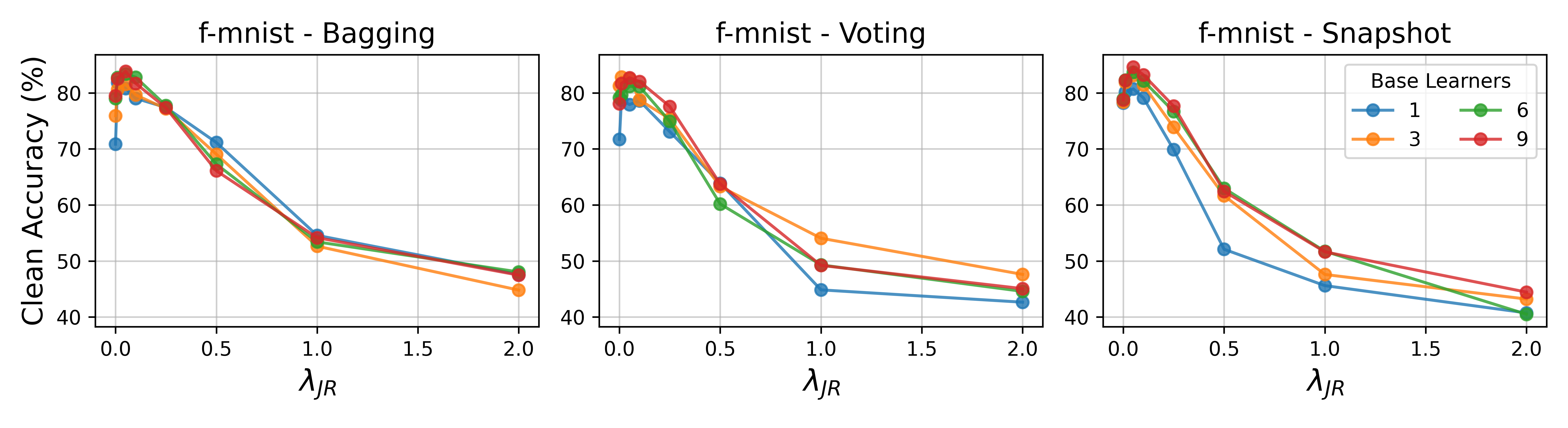}
\caption{Clean accuracy of LeNet on MNIST (top) and Fashion-MNIST (bottom) for various Jacobian regularization strengths $\lambda_{\text{JR}}$ and with varying number of base learners per ensemble method.}
\label{fig:testacc}
\end{figure}

\begin{figure}[ht]
\includegraphics[width=\textwidth]{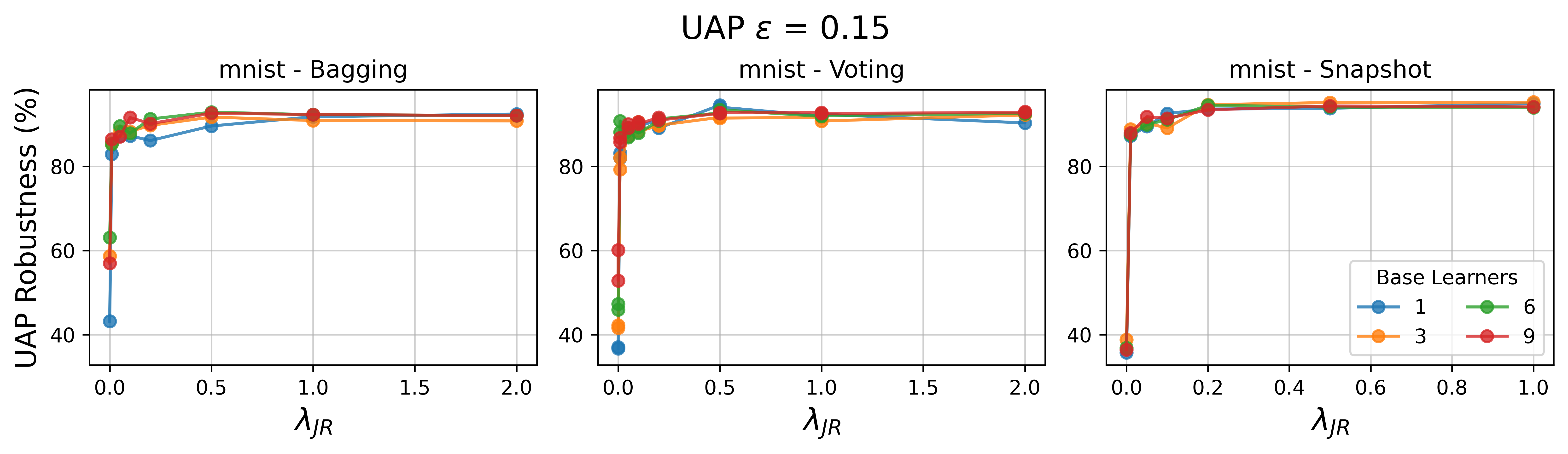}
\includegraphics[width=\textwidth]{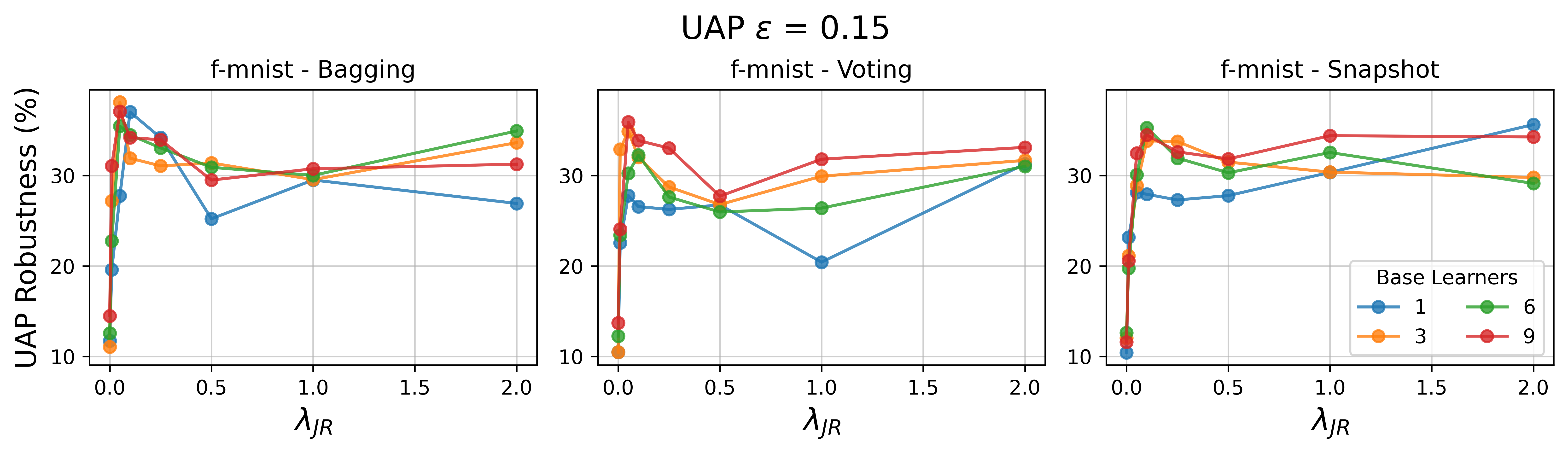}
\caption{Model robustness against UAP with $\varepsilon = 0.15$ of LeNet on MNIST (top) and Fashion-MNIST (bottom) for various Jacobian regularization strengths $\lambda_{\text{JR}}$ and with varying number of base learners per ensemble method.}
\label{fig:uer}
\end{figure}

\vspace{3pt}\noindent\textbf{Metrics.} The following metrics are evaluated on the entire 10,000 sample test sets for each dataset. \emph{Clean Accuracy} is the accuracy of the model on the test set. \emph{Model Robustness} measures the accuracy of the model on the test set when the corresponding worst-case UAP is applied or present. We then average this model robustness over all the UAP attack scenarios that we consider, $\ell_{\infty}$-norm of $\varepsilon = 0.10, 0.15, 0.2, 0.25$, to get an overall \emph{mean UAP Accuracy}.

\subsection{Improvements with Jacobian Ensembles}

\textbf{Clean Accuracy.} In Fig.~\ref{fig:testacc} there is a rapid degradation in clean accuracy when $\lambda_{\text{JR}}$ is large. This is when JR is more heavily weighted, and this is consistent with previous work \cite{co2021jacobian,hoffman2019robust} as too much regularization damages accuracy on the test set. However, having a small amount of JR is still noticeably more beneficial than no JR as indicated by the clean accuracy when considering $\lambda_{\text{JR}}$ in the range of 0 to 0.1 for both datasets across all settings.

We also see benefits of using ensembles: models with more than one base learner have noticeably better clean accuracy across all settings in Fig.~\ref{fig:testacc}. Overall, Jacobian Ensembles, which is a combination of ensemble methods and JR, achieves the best performance in our experiments.

\begin{figure}[ht]
\includegraphics[width=\textwidth]{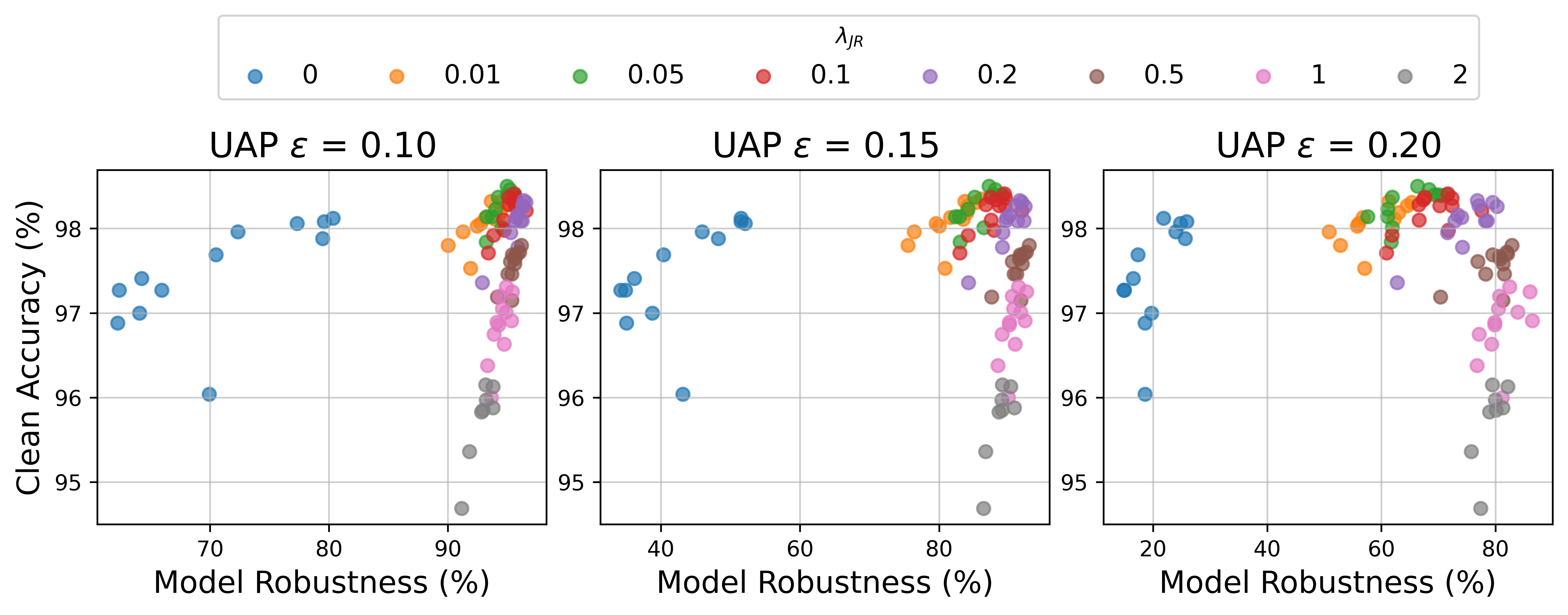}
\includegraphics[width=\textwidth]{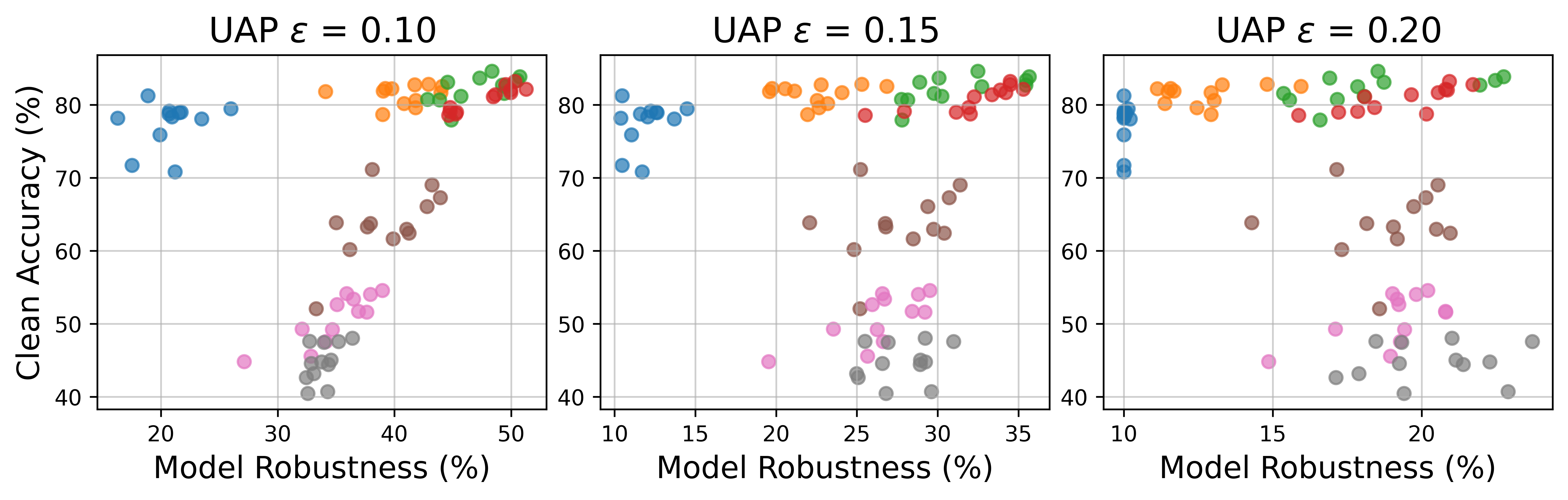}\\
\caption{Clean accuracy versus robustness trade-off of LeNet on MNIST (top) and Fashion-MNIST (bottom) labeled by $\lambda_{\text{JR}}$ for various UAP attack strengths $\varepsilon$.}
\label{fig:tradeoff}
\end{figure}

\begin{table}[ht]
\small
\centering
\caption{Weighted accuracy (in \%) of LeNet for MNIST (top) and Fashion-MNIST (bottom). The first 3 rows show the models with highest weighted accuracy. The bottom 3 rows show the \emph{best} weighted accuracy of models trained with only JR, only ensemble, and ``standard training'' (neither JR nor ensemble).}
\begin{tabular}{rcc|ccc}
    \multicolumn{3}{c}{Model} & \multicolumn{3}{c}{\hspace{1mm} MNIST Accuracy (\%)} \\
    \hspace{1mm} Ensemble & \hspace{1mm} Learners & \hspace{1mm} $\lambda_{\text{JR}}$ & \hspace{1mm} Clean & \hspace{1mm} Avg. UAP & \hspace{1mm} \textbf{Weighted}\\
	\hline
	\hline
	Snapshot & \hspace{1mm} 3 & \hspace{1mm} 0.50 \hspace{1mm} & \hspace{1mm} 97.8 & \hspace{1mm} 82.8 & \hspace{1mm} \textbf{90.3}\\
	Snapshot & \hspace{1mm} 9 & \hspace{1mm} 0.50 \hspace{1mm} & \hspace{1mm} 97.7 & \hspace{1mm} 82.3 & \hspace{1mm} 90.0\\
	Snapshot & \hspace{1mm} 3 & \hspace{1mm} 0.20 \hspace{1mm} & \hspace{1mm} 98.3 & \hspace{1mm} 81.5 & \hspace{1mm} 89.9\\
	\hline
	JR Only & \hspace{1mm} 1 & \hspace{1mm} 0.05 \hspace{1mm} & \hspace{1mm} 98.4 & \hspace{1mm} 74.2 & \hspace{1mm} 86.3\\
	Bagging & \hspace{1mm} 6 & \hspace{1mm} 0 \hspace{1mm} & \hspace{1mm} 98.1 & \hspace{1mm} 42.8 & \hspace{1mm} 70.4\\
	Standard & \hspace{1mm} 1 & \hspace{1mm} 0 \hspace{1mm} & \hspace{1mm} 99.1 & \hspace{1mm} 31.9 & \hspace{1mm} 65.5\\
\end{tabular}
\\
\vspace{2.5mm}
\begin{tabular}{rcc|ccc}
    \multicolumn{3}{c}{Model} & \multicolumn{3}{c}{\hspace{1mm} Fashion-MNIST Accuracy (\%)} \\
    \hspace{1mm} Ensemble & \hspace{1mm} Learners & \hspace{1mm} $\lambda_{\text{JR}}$ & \hspace{1mm} Clean & \hspace{1mm} Avg. UAP & \hspace{1mm} \textbf{Weighted}\\
	\hline
	\hline
	Bagging & \hspace{1mm} 9 & \hspace{1mm} 0.05 \hspace{1mm} & \hspace{1mm} 83.9 & \hspace{1mm} 43.2 & \hspace{1mm} \textbf{63.5}\\
	Bagging & \hspace{1mm} 6 & \hspace{1mm} 0.05 \hspace{1mm} & \hspace{1mm} 83.4 & \hspace{1mm} 43.0 & \hspace{1mm} 63.2\\
	Snapshot & \hspace{1mm} 9 & \hspace{1mm} 0.10 \hspace{1mm} & \hspace{1mm} 83.2 & \hspace{1mm} 42.4 & \hspace{1mm} 62.8\\
	\hline
	JR Only & \hspace{1mm} 1 & \hspace{1mm} 0.10 \hspace{1mm} & \hspace{1mm} 79.0 & \hspace{1mm} 31.2 & \hspace{1mm} 55.1\\
	Bagging & \hspace{1mm} 9 & \hspace{1mm} 0 \hspace{1mm} & \hspace{1mm} 91.2 & \hspace{1mm} 12.7 & \hspace{1mm} 51.2\\
	Standard & \hspace{1mm} 1 & \hspace{1mm} 0 \hspace{1mm} & \hspace{1mm} 90.8 & \hspace{1mm} 12.1 & \hspace{1mm} 50.9\\
\end{tabular}
\label{table:weighted}
\end{table}

\vspace{3pt}\noindent\textbf{Model Robustness.} In the interest of space, we only present the robustness results for a particular UAP attack strength $\varepsilon = 0.15$ in Fig.~\ref{fig:uer}. In terms of robustness, trends for both datasets are slightly different since Fashion-MNIST is a more difficult dataset than MNIST: a regularly trained LeNet on MNIST can be expected to have 98-99\% clean accuracy whereas it is 90-91\% for the same model on Fashion-MNIST. Thus, it is expected that models on Fashion-MNIST are considerably less robust. For most settings, the general trends in Fig.~\ref{fig:uer} show that ensembles have a robustness benefit. JR monotonically increases robustness for MNIST and for Fashion-MNIST has a range between $0 < \lambda_{\text{JR}} < 0.5$ that achieves the best robustness.

In conclusion, ensemble methods with more than one base learner introduce measurable advantages in both accuracy and robustness. These advantages become even more pronounced when combined with JR. Next we analyze the general trade-off between accuracy and robustness.

\subsection{Accuracy-Robustness Trade-Off}
In Fig.~\ref{fig:tradeoff}, we plot the accuracy versus robustness of the various models we trained under the different configurations accounting for various ensemble methods, number of base learners, and $\lambda_{\text{JR}}$ values. The best models achieve both high accuracy and robustness, so these will be on the top right side of the graph.

In a scenario when robustness is not accounted for (i.e. $\lambda_{\text{JR}} = 0$), models would appear in the top left. They are trained to have good accuracy, but remain extremely vulnerable to adversarial attacks like UAPs. On the other hand, models that overcompensate for robustness such as those with very high regularization (e.g. $\lambda_{\text{JR}} = 2$), will appear on the bottom right. As these models sacrifice a significant amount of clean accuracy for improved robustness, especially against UAP attacks with larger strength $\varepsilon$. These delineations become clear when labeling the models according to their $\lambda_{\text{JR}}$ value as in Fig.~\ref{fig:tradeoff}. Thus, the role of $\lambda_{\text{JR}}$ is evident in improving overall robustness.

To better capture the model performance on both benign and adversarial inputs, we compute the \emph{Weighted Accuracy} by averaging the clean accuracy and mean UAP accuracy. In practice, the defender can adjust the weighting of each accuracy metric in their final assessment to better match their application and risk profile. We choose the mean as the base setting.

Next, we perform an ablation study on the effect of only JR and only ensemble models compared against the top 3 Jacobian Ensembles with the best weighted accuracy in Table~\ref{table:weighted}. We find that Jacobian Ensembles achieve the best weighted accuracy. To compare, we also show in the bottom 3 rows for each table the best models with only JR, only ensembles, and neither JR nor ensembles. The best Jacobian Ensembles have a clear advantage with average UAP accuracy over the the non-Jacobian Ensembles whilst maintaining very close clean accuracy. This difference is even more pronounced on the Fashion-MNIST dataset.

Differences between the two datasets MNIST and Fashion-MNIST also show in Table~\ref{table:weighted}. Since MNIST is an easier dataset, performance degradation by large $\lambda_{\text{JR}} = 0$ are not as prominent, so larger $\lambda_{\text{JR}}$ are favored by the combined score. For Fashion-MNIST, a large $\lambda_{\text{JR}}$ is detrimental as the base model begin with a relatively low clean accuracy ($< 91$\%). In both cases, ensembles demonstrate a noticeably large boost in weighted accuracy, and further tuning is likely to improve their performance.

\section{Conclusion}
In this work, we propose Jacobian Ensembles to significantly increase model robustness against UAPs whilst maintaining the clean accuracy of models. Our results show that Jacobian Ensembles takes the advantages of both Jacobian regularization and model ensembles to achieve superior accuracy and robustness than each of these methods on their own, as measured by our weighted metric.

In addition, we derive theoretical upper and lower bounds on the robustness to UAPs for model ensembles, showing that increasing the number of base classifiers in the models' ensembles reduces the expected Frobenius norm of their Jacobian and thus improves stability. We then empirically verify our results and show that a combination of both JR and ensembles achieve the best performance.

These results give us confidence in recommending Jacobian Ensembles as a general methodology when training models as UAPs present a great threat to model adoption and safety. Our results show that it is indeed possible to maintain great test accuracy whilst achieving significant UAP robustness in previously unseen levels of accuracy-robustness trade-off. Thus, it is indeed possible to get the best of both worlds.

\bibliographystyle{splncs04}
\bibliography{main}

\begin{thebibliography}{10}
\providecommand{\url}[1]{\texttt{#1}}
\providecommand{\urlprefix}{URL }
\providecommand{\doi}[1]{https://doi.org/#1}

\bibitem{breiman1996bagging}
Breiman, L.: Bagging predictors. Machine learning  \textbf{24}(2),  123--140
  (1996)

\bibitem{brown2017adversarial}
Brown, T.B., Man{\'e}, D.: Adversarial patch. arXiv preprint arXiv:1712.09665
  (2017)

\bibitem{co2019procedural}
Co, K.T., Mu\~{n}oz Gonz\'{a}lez, L., de~Maupeou, S., Lupu, E.C.: Procedural
  noise adversarial examples for black-box attacks on deep convolutional
  networks. In: Proceedings of the 2019 ACM SIGSAC Conf. on Computer and
  Communications Security. pp. 275--289. CCS '19 (2019).
  \doi{10.1145/3319535.3345660}

\bibitem{co2019universal}
Co, K.T., Mu{\~n}oz-Gonz{\'a}lez, L., Kanthan, L., Glocker, B., Lupu, E.C.:
  Universal adversarial robustness of texture and shape-biased models. arXiv
  preprint arXiv:1911.10364  (2019)

\bibitem{co2019sensitivity}
Co, K.T., Mu{\~n}oz-Gonz{\'a}lez, L., Lupu, E.C.: Sensitivity of deep
  convolutional networks to gabor noise. arXiv preprint arXiv:1906.03455
  (2019)

\bibitem{co2021jacobian}
Co, K.T., Rego, D.M., Lupu, E.C.: Jacobian regularization for mitigating
  universal adversarial perturbations. In: International Conference on
  Artificial Neural Networks. pp. 202--213. Springer (2021)

\bibitem{eykholt2018physical}
Eykholt, K., Evtimov, I., Fernandes, E., Li, B., Rahmati, A., Tramer, F.,
  Prakash, A., Kohno, T., Song, D.: Physical adversarial examples for object
  detectors. In: 12th {USENIX} Workshop on Offensive Technologies ({$WOOT$} 18)
  (2018)

\bibitem{eykholt2018robust}
Eykholt, K., Evtimov, I., Fernandes, E., Li, B., Rahmati, A., Xiao, C.,
  Prakash, A., Kohno, T., Song, D.: Robust physical-world attacks on deep
  learning visual classification. In: Proceedings of the IEEE Conf. on Computer
  Vision and Pattern Recognition (CVPR). pp. 1625--1634 (2018)

\bibitem{freund1999short}
Freund, Y., Schapire, R., Abe, N.: A short introduction to boosting.
  Journal-Japanese Society For Artificial Intelligence  \textbf{14}(771-780),
  ~1612 (1999)

\bibitem{hau2021object}
Hau, Z., Co, K.T., Demetriou, S., Lupu, E.C.: Object removal attacks on
  lidar-based 3d object detectors. arXiv preprint arXiv:2102.03722  (2021)

\bibitem{hau2021shadow}
Hau, Z., Demetriou, S., Mu{\~n}oz-Gonz{\'a}lez, L., Lupu, E.C.: Shadow-catcher:
  Looking into shadows to detect ghost objects in autonomous vehicle 3d
  sensing. In: Euro. Symposium on Research in Computer Security. pp. 691--711.
  Springer (2021)

\bibitem{hinton2012deep}
Hinton, G., Deng, L., Yu, D., Dahl, G.E., Mohamed, A.r., Jaitly, N., Senior,
  A., Vanhoucke, V., Nguyen, P., Sainath, T.N., et~al.: Deep neural networks
  for acoustic modeling in speech recognition: The shared views of four
  research groups. IEEE Signal Processing Magazine  \textbf{29}(6),  82--97
  (2012)

\bibitem{hoffman2019robust}
Hoffman, J., Roberts, D.A., Yaida, S.: Robust learning with jacobian
  regularization. arXiv preprint arXiv:1908.02729  (2019)

\bibitem{huang2017snapshot}
Huang, G., Li, Y., Pleiss, G., Liu, Z., Hopcroft, J.E., Weinberger, K.Q.:
  Snapshot ensembles: Train 1, get m for free. In: Intl. Conf. on Learning Rep.
  (2017)

\bibitem{krizhevsky2012imagenet}
Krizhevsky, A., Sutskever, I., Hinton, G.E.: Imagenet classification with deep
  convolutional neural networks. In: Advances in Neural Information Processing
  Systems (NeurIPS). pp. 1097--1105 (2012)

\bibitem{kuncheva2014combining}
Kuncheva, L.I.: Combining pattern classifiers: methods and algorithms. John
  Wiley \& Sons (2014)

\bibitem{lecun1998gradient}
LeCun, Y., Bottou, L., Bengio, Y., Haffner, P.: Gradient-based learning applied
  to document recognition. Proceedings of the IEEE  \textbf{86}(11),
  2278--2324 (1998)

\bibitem{matachana2020robustness}
Matachana, A.G., Co, K.T., Mu{\~n}oz-Gonz{\'a}lez, L., Martinez, D., Lupu,
  E.C.: Robustness and transferability of universal attacks on compressed
  models. arXiv preprint arXiv:2012.06024  (2020)

\bibitem{moosavi2017universal}
Moosavi-Dezfooli, S.M., Fawzi, A., Fawzi, O., Frossard, P.: Universal
  adversarial perturbations. In: Proceedings of the IEEE Conf. on Computer
  Vision and Pattern Recognition (CVPR). pp. 1765--1773 (2017)

\bibitem{redmon2016you}
Redmon, J., Divvala, S., Girshick, R., Farhadi, A.: You only look once:
  Unified, real-time object detection. In: Proceedings of the IEEE Conf. on
  Computer Vision and Pattern Recognition (CVPR). pp. 779--788 (2016)

\bibitem{roth2020adversarial}
Roth, K., Kilcher, Y., Hofmann, T.: Adversarial training is a form of
  data-dependent operator norm regularization. In: Advances in Neural
  Information Processing Systems (NeurIPS) (2020)

\bibitem{shafahi2018universal}
Shafahi, A., Najibi, M., Xu, Z., Dickerson, J., Davis, L.S., Goldstein, T.:
  Universal adversarial training. arXiv preprint arXiv:1811.11304  (2018)

\bibitem{thys2019fooling}
Thys, S., Van~Ranst, W., Goedem\'e, T.: Fooling automated surveillance cameras:
  adversarial patches to attack person detection. In: CVPRW: Workshop on The
  Bright and Dark Sides of Computer Vision: Challenges and Opportunities for
  Privacy and Security (2019)

\bibitem{tramer2019adversarial}
Tram\`{e}r, F., Dupr\'{e}, P., Rusak, G., Pellegrino, G., Boneh, D.:
  Adversarial: Perceptual ad blocking meets adversarial machine learning. In:
  Proceedings of the 2019 ACM SIGSAC Conf. on Computer and Communications
  Security. p. 2005–2021. CCS ’19 (2019). \doi{10.1145/3319535.3354222}

\bibitem{xiao2017fashion}
Xiao, H., Rasul, K., Vollgraf, R.: Fashion-mnist: a novel image dataset for
  benchmarking machine learning algorithms. arXiv preprint arXiv:1708.07747
  (2017)

\bibitem{zhou2012ensemble}
Zhou, Z.H.: Ensemble methods: foundations and algorithms. CRC press (2012)

\end{thebibliography}

\end{document}